\documentclass[twoside]{article}

\usepackage{amsmath}
\usepackage{amsthm}
\usepackage{amssymb}
\usepackage{hyperref}
\usepackage{graphicx}
\usepackage{color}
\usepackage[usenames]{xcolor}
\usepackage{rotating}
\usepackage{enumerate,enumitem}
\usepackage{verbatim}
\usepackage{dsfont}

\usepackage{tikz}

\newtheorem{lemma}{Lemma}
\newtheorem{thm}[lemma]{Theorem}

\theoremstyle{definition}

\newcommand{\Ecal}{\ensuremath{\mathcal{E}}}
\newcommand{\Mcal}{\ensuremath{\mathcal{M}}}
\newcommand{\Scal}{\ensuremath{\mathcal{S}}}
\newcommand{\Xcal}{\ensuremath{\mathcal{X}}}
\newcommand{\Ycal}{\ensuremath{\mathcal{Y}}}

\newcommand{\indmod}{\Mcal_1}
\newcommand{\Dir}{\operatorname{Dir}}

\newcommand{\<}{\ensuremath{\langle}}
\renewcommand{\>}{\ensuremath{\rangle}}

\newcommand{\rd}{\ensuremath{\mathrm{d}}}

\DeclareMathOperator*{\arginf}{arginf}
\DeclareMathOperator{\supp}{supp}

\newcommand{\ba}{\boldsymbol{\alpha}}
\newcommand{\al}{\alpha}

\title{\sc Scaling of Model Approximation Errors\\and Expected Entropy Distances}
\author{
  Guido F. Mont\'ufar\\
     {\small Department of Mathematics} \\
     {\small Pennsylvania State University} \\
     {\small University Park PA 16802 USA} \\
     {\small e-mail: gfm10@psu.edu}
  \and
  Johannes Rauh\\
     {\small Max Planck Institute}\\ {\small for Mathematics in the Sciences} \\
     {\small Inselstr. 22 04103 Leipzig Germany} \\
     {\small e-mail: jrauh@mis.mpg.de}
  } 

\newfont{\m}{cmr8}
\newfont{\ms}{cmsl8}
\begin{document}
\maketitle

\begin{abstract}
  We compute the expected value of the Kullback-Leibler divergence to various fundamental statistical models with
  respect to canonical priors on the probability simplex. We obtain closed formulas for the expected model
  approximation errors, depending on the dimension of the models and the cardinalities of their sample spaces. 
  For the uniform prior, the expected divergence from any model containing the uniform distribution is bounded by a
  constant~$1-\gamma$, and for the models that we consider, this bound is approached if the state space is very large and
  the models' dimension does not grow too fast.  For Dirichlet priors the expected divergence is bounded in a similar way, if the concentration parameters take reasonable values. 
  These results serve as reference values for more complicated statistical models. 
\end{abstract}

\section{Introduction}

\thispagestyle{empty}

Let $p,q$ be probability distributions on a finite set~$\Xcal$.
The \emph{information divergence}, \emph{relative entropy}, or \emph{Kullback-Leibler divergence}
\begin{equation*}
D(p\|q) = \sum_{i\in\Xcal}p_{i}\log\frac{p_{i}}{q_{i}}
\end{equation*}
is a natural measure of dissimilarity between $p$ and $q$. It specifies how easily the two distributions can be distinguished from each other by means of statistical experiments.  In this paper we use the natural logarithm.
The divergence is related to the log-likelihood: If $p$ is an empirical distribution, summarizing the outcome of $n$
statistical experiments, then the log-likelihood of a distribution $q$ equals $- n (D(p\|q) + H(p))$, where $H(p)$ is the Shannon entropy of $p$. 
Hence finding a \emph{maximum likelihood estimate} $q$ within a set of probability distributions $\Mcal$ is the same as
finding a minimizer of the divergence $D(p\|q)$ with $q$ restricted to $\Mcal$. 

Assume that $\Mcal^{\text{true}}$ is a set of probability distributions for which there is no simple mathematical description available. 
We would like to identify a model $\Mcal$ which does not necessarily contain all distributions from $\Mcal^{\text{true}}$, but which approximates them relatively well. What error magnitude should we accept from a good model?  

To assess the expressive power of a model $\Mcal$, we study the function $p\mapsto D(p\|\Mcal) = \inf_{q\in\Mcal}
D(p\|q)$.  Finding the maximizers of this function corresponds to a worst-case analysis.  The problem of maximizing the
divergence from a statistical model was first posed in~\cite{Ay02:Pragmatic_structuring}, motivated by infomax principles in the context of neural networks.  Since then, important progress has been made,
especially in the case of exponential
families~\cite{MatusRauh11:Maximization-ISIT2011,MatusAy03:On_Maximization_of_the_Information_Divergence,Rauh11:Thesis},
but also in the case of discrete mixture models and restricted Boltzmann machines~\cite{NIPS2011_0307}. 

In addition to the worst-case error bound, the expected performance and \emph{expected error} are of interest. 
This leads to the mathematical problem of computing the expectation value
\begin{equation}
  \langle D(p\|\Mcal) \rangle = \int_{\Delta} D(p\|\Mcal)\, \psi(p) \,\rd p\;,\label{integral}
\end{equation}
where $p$ is drawn from a {\em prior} probability density $\psi$ on the probability simplex~$\Delta$. 
The correct prior depends on the concrete problem at hand and is often difficult to determine.  
We ask: Given conditions on the prior, how different is the worst case from the average case? To what extent can both errors be influenced by the choice of the model? 
We focus on the case of Dirichlet priors. 
It turns out that in most cases the worst-case error diverges as the number of elementary events $N$ tends to infinity, while the expected error remains bounded. 
Our analysis leads to integrals that have been considered in 
Bayesian function estimation in~\cite{WolpertWolf95:Estimating_functions_of_probability_distributions}, and we can take advantage of the tools developed there.

Our first observation is that, if $\psi$ is the uniform prior, then the expected divergence from $p$ to the uniform distribution
is a monotone function of the system size $N$ and converges to the constant
$1-\gamma\approx0.4228$ 
as $N\to\infty$, where $\gamma$ is the \emph{Euler-Mascheroni} constant. 
Many natural statistical models contain the uniform distribution; and the expected divergence from such models is
bounded by the same constant. 
On the other hand, when $p$ and $q$ are chosen uniformly at random, the expected divergence $\langle D(p\|q)\rangle_{p,q}$ is equal to $1-1/N$. 

We show, for a class of models including independence models, partition models, mixtures of product distributions with disjoint
supports~\cite{NIPS2011_0307}, and decomposable hierarchical models, that the expected divergence actually has the same limit, $1-\gamma$, provided the dimension of the models remains \emph{small} with respect to~$N$ (the usual case in applications). 
For Dirichlet priors the results are similar (for reasonable choices of parameters). 
In contrast, when $\Mcal$ is an exponential family, the maximum value of $D(\cdot\|\Mcal)$ is at least $\log(N/(\operatorname{dim}(\Mcal) + 1))$, see~\cite{Rauh13:Optimal_Expfams}. 

In Section~\ref{sec:prelim} we define various model classes and collect basic properties of Dirichlet priors. 
Section~\ref{sec:evalues} contains our main results: closed-form expressions for the expectation values of entropies and divergences. 
The results are discussed in Section~\ref{sec:interpretation}.  
Proofs and calculations are deferred to Appendix~\ref{sec:proofs}.

\section{Preliminaries}
\label{sec:prelim}

\subsection{Models from statistics and machine learning}
\label{ssec:models}

We consider random variables on a finite set of elementary events $\Xcal$, $|\Xcal|=N$. 
The set of probability distributions on $\Xcal$ is the $(N-1)$-simplex $\Delta_{N-1}\subset\mathbb{R}^N$. 
A \emph{model} is a subset of~$\Delta_{N-1}$. 
The support sets of a model $\Mcal\subseteq\Delta_{N-1}$ are the support sets $\supp(p)=\{i\in\Xcal\,|\, p_{i}>0\}$ of points $p=(p_i)_{i\in\Xcal}$ in $\Mcal$. 

The {\em $K$-mixture} of a model $\Mcal$ is the union of all convex combinations of any $K$ of its points: $\Mcal^K:=\{\sum_{i=1}^K \lambda_i p^{(i)}\,|\, \lambda_i\geq0, \sum_i\lambda_i=1, p^{(i)}\in\Mcal\}$. 
Given a partition $\varrho=\{A_1,\ldots,A_K\}$ of $\Xcal$ into $K$ disjoint support sets of $\Mcal$, the {\em $K$-mixture of $\Mcal$ with disjoint supports $\varrho$} is the subset of $\Mcal^K$ defined by
\begin{equation*}
  \Mcal^\varrho=\left\{\sum_{i=1}^K \lambda_i p^{(i)}\in\Mcal^{K}\,\middle|\, p^{(i)}\in\Mcal, \supp(p^{(i)})\subseteq A_i \text{ for all }i\right\}.
\end{equation*}

Let $\varrho=\{A_1,\ldots, A_K\}$ be a partition of $\Xcal$.  The {\em partition model} $\Mcal_\varrho$ consists of all
$p\in\Delta_{N-1}$ that satisfy $p_{i} = p_{j}$ whenever $i,j$ belong to the same block in the partition $\varrho$.  Partition models
are closures of convex exponential families with uniform reference measures.  The closure of a convex
exponential family is a set of the form (see~\cite{MatusAy03:On_Maximization_of_the_Information_Divergence})
\begin{equation*}
  \Mcal_{\varrho,\nu}=\left\{\sum_{k=1}^K \lambda_k \frac{ \mathds{1}_{A_k} \nu}{ \nu(A_k)}  \,\middle|\, \lambda_k\geq 0,\sum_{k=1}^K \lambda_k=1\right\},
\end{equation*}
where $\nu:\Xcal\to(0,\infty)$ is a positive function on $\Xcal$ called \emph{reference measure}, and $\mathds{1}_A$ is the indicator function of $A$. 
Note that all measures $\nu$ with fixed conditional distributions $\nu(\cdot|A_k)={\nu(\cdot)}/{\sum_{j\in A_k}\nu(j)}$ on $A_k$, for all $k$, yield the same model.  In
fact, $\Mcal_{\varrho,\nu}$ is the $K$-mixture of the set $\{\nu(\cdot|A_{k}) : k=1,\dots,K \}$.

For a composite system with $n$ variables $X_1,\ldots,X_n$, the set of elementary events is $\Xcal=\Xcal_1\times\cdots\times\Xcal_n$, $|\Xcal_i|=N_i$ for all~$i$.
A \emph{product distribution} is a distribution of the form
\begin{equation*}
  p(x_{1},\dots,x_{n}) = p_{\{1\}}(x_{1})\cdots p_{\{n\}}(x_{n})\qquad\text{for all } x\in\Xcal,
\end{equation*}
where $p_{\{i\}}\in\Delta_{N_{i}-1}$. The \emph{independence model} $\Mcal_1$ is the set of all product distributions on~$\Xcal$. 
The support sets of the independence model are the sets of the form $A=\Ycal_1\times\cdots\times\Ycal_n$ with
$\Ycal_i\subseteq\Xcal_i$ for each $i$.

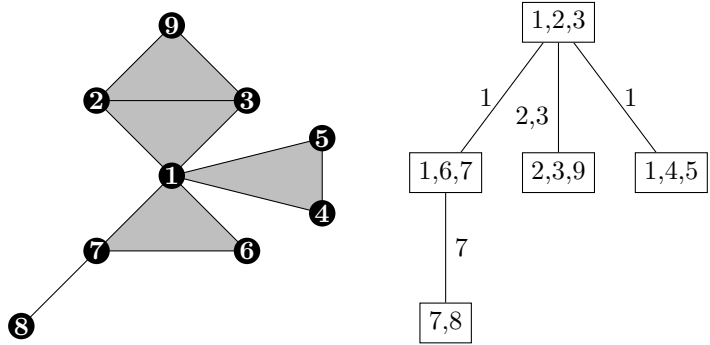
\begin{figure}
  \centering
  \begin{tikzpicture}
    \path (2,2) coordinate (X1);
    \path (1,3) coordinate (X2);
    \path (3,3) coordinate (X3);
    \path (4,1.5) coordinate (X4);
    \path (4,2.5) coordinate (X5);
    \path (3,1) coordinate (X6);
    \path (1,1) coordinate (X7);
    \path (0,0) coordinate (X8);
    \path (2,4) coordinate (X9);
    \begin{scope}[fill=lightgray]
      \filldraw (X2) -- (X9) -- (X3);
      \filldraw (X1) -- (X2) -- (X3) -- cycle;
      \filldraw (X1) -- (X4) -- (X5) -- cycle;
      \filldraw (X1) -- (X6) -- (X7) -- cycle;
      \draw (X7) -- (X8);
    \end{scope}
    \foreach \i in {1,...,9} { \fill (X\i) circle (5pt); }
    \foreach \i in {1,...,9} { \draw[white] (X\i) node {\textbf{\i}}; }
  \end{tikzpicture}
  $\qquad$
  \begin{tikzpicture}
    \begin{scope}[shape=rectangle,every node/.style={draw}]
      \path (1.5,4) node (123) {1,2,3};
      \path (0.0,2) node (167) {1,6,7};
      \path (1.5,2) node (239) {2,3,9};
      \path (3.0,2) node (145) {1,4,5};
      \path (0.0,0) node (78)  {7,8};
    \end{scope}
    \draw (78) to node[right] {7} (167) to node[left] {1} (123) to node[right] {1} (145);
    \draw (123) to node[below left] {2,3} (239);
  \end{tikzpicture}
  \caption{An example of a decomposable model and its junction tree.}
  \label{fig:decomposable-model}
\end{figure}

Let $\Scal$ be a simplicial complex on $\{1,\dots,n\}$.  The \emph{hierarchical model} $\Mcal_{\Scal}$ consists of all
probability distributions that have a factorization of the form $p(x) = \prod_{S\in\Scal}\Phi_{S}(x)$, where $\Phi_{S}$
is a positive function that depends only on the $S$-coordinates of $x$.  The model $\Mcal_{\Scal}$ is called
\emph{reducible} if there exist simplicial subcomplexes $\Scal_{1},\Scal_{2}\subset\Scal$ such that
$\Scal_{1}\cup\Scal_{2}=\Scal$ and $\Scal_{1}\cap\Scal_{2}$ is a simplex.  In this case, the set
$(\bigcup_{\Ycal\in\Scal_{1}}\Ycal)\cap(\bigcup_{\Ycal\in\Scal_{2}}\Ycal)$ is called a \emph{separator}. Furthermore, $\Mcal_{\Scal}$ is \emph{decomposable} if it can be iteratively reduced into simplices.  Such an iterative reduction can
be described by a \emph{junction tree}, which is a tree $(V,E)$ with vertex set the set of facets of $\Scal$ and with
edge labels the separators. 
The independence model is an example of a decomposable model. 
We give another example in Fig.~\ref{fig:decomposable-model} and refer to~\cite{DrtonSturmfelsSullivant09:Algebraic_Statistics} for more details. 
In general, the junction tree is not unique, but the multi-set of separators is unique. 

\medskip

For most models there is no closed-form expression for $D(\cdot\|\Mcal)$, since there is no closed formula for
$\arginf_{q\in\Mcal} D(p\|q)$.  However, for some of the models mentioned above a closed formula does exist:

The divergence from the independence model is called {\em multi-information} and satisfies
\begin{equation}
  \label{eq:MI}
  MI(X_{1},\dots,X_{n}) = D(p\|\indmod) = -H(X_1,\ldots,X_n)  + \sum_{k=1}^n H(X_k).
\end{equation}
If $n=2$ it is also called the \emph{mutual information} of $X_{1}$ and $X_{2}$. 
The divergence from $\Mcal_{\varrho,\nu}$ equals (see~\cite[eq.~(1)]{MatusAy03:On_Maximization_of_the_Information_Divergence})
\begin{equation}
  \label{eq:div-from-conv}
  D(p\|\Mcal_{\varrho,\nu}) = D(p\|\sum_{k=1}^K p(A_k) \nu(x|A_k)).
\end{equation}
For a decomposable model $\Mcal_{\Scal}$ with junction tree $(V,E)$,
\begin{equation}
  \label{eq:decomp-div}
  D(p\|\Mcal_{\Scal}) = \sum_{S\in V} H_{p}(X_{S}) - \sum_{S\in E} H_{p}(X_{S}) - H(p).
\end{equation}
Here, $H_{p}(X_{S})$ denotes the joint entropy of the random variables $\{X_{i}\}_{i\in S}$ under~$p$.

\subsection{Dirichlet prior}
\label{ssec:dirichlet}

The Dirichlet distribution (or Dirichlet prior) with {\em concentration parameter} $\boldsymbol{\alpha}=(\alpha_1,\ldots,\alpha_N)\in\mathbb{R}_{>0}^N$, is the probability distribution on $\Delta_{N-1}$ with density $\Dir_{\ba}(p) :=  \frac{1}{\sqrt{N}}\frac{\Gamma(\sum_{i=1}^N\alpha_i) }{\prod_{i=1}^N\Gamma(\alpha_i)} \prod_{i=1}^N p_i^{\alpha_i-1}$ for all $p=(p_1,\ldots,p_N)\in\Delta_{N-1}$, where $\Gamma$ is the gamma function. 
We write $\alpha=\sum_{i=1}^N\alpha_i$. 

Note that $\Dir_{(1,\ldots,1)}$ is the uniform probability density on $\Delta_{N-1}$.  Furthermore, note that $\lim_{a\to0}\Dir_{(a,\ldots,a)}$ is uniformly concentrated in the point measures (it assigns mass $1/N$ to $\delta_x$, for all $x\in\Xcal$), and  $\lim_{a\to\infty}\Dir_{(a,\ldots,a)}$ is concentrated in the uniform distribution $u:=(1/N,\ldots,1/N)$. In general, if $\ba\in\Delta_{N-1}$, then $\lim_{\kappa\to\infty}\Dir_{\kappa\ba}$ is the Dirac delta concentrated on $\ba$.  

A basic property of the Dirichlet distributions is the {\em aggregation property}:   
Consider  a partition $\varrho=\{A_1,\ldots, A_K\}$  of $\Xcal=\{1,\ldots,N\}$. 
If $p=(p_1,\ldots,p_N)\sim\Dir_{(\alpha_1,\ldots,\alpha_N)}$, then $(\sum_{i\in A_1}p_i, \ldots, \sum_{i\in A_K} p_i)\sim\Dir_{(\sum_{i\in A_1}\alpha_i, \ldots, \sum_{i\in A_K}\alpha_i)}$, see, e.g.,~\cite{DirIntro}. 
We write $\ba^\varrho=(\al^\varrho_1,\ldots,\al^\varrho_K)$, $\al^\varrho_k=\sum_{i\in A_k}\al_i$ for the concentration parameter induced by the partition $\varrho$. 

The aggregation property is useful when treating marginals of composite systems. 
Given a composite system with $\Xcal=\Xcal_1\times\cdots\times\Xcal_n$, $|\Xcal|=N$, $\Xcal_k=\{1,\ldots,N_k\}$, 
we write ${\boldsymbol\alpha}^k=(\alpha^k_1,\ldots, \alpha^k_{N_k})$,  $\alpha^k_j=\sum_{x\in\Xcal\colon x_k=j}\alpha_x$ for the concentration parameter of the Dirichlet distribution induced on the $\Xcal_k$-marginal $(\sum_{x\in\Xcal\colon x_k=1}p(x),\ldots, \sum_{x\in\Xcal\colon x_k=N_k}p(x))$.

\section{Expected entropies and divergences}
\label{sec:evalues}

For any $k\in\mathbb{N}$ let $h(k) = 1 + \frac12 + \dots + \frac1k$ be the $k$th \emph{harmonic number}.  It is known that for large $k$,
\begin{equation*}
  h(k) = \log(k) + \gamma + O(\tfrac1k),
\end{equation*}
where $\gamma\approx0.57721$ is the \emph{Euler-Mascheroni constant}.  Moreover, $h(k) - \log(k)$ is strictly positive
and decreases monotonically.  We also need the natural analytic extension of $h$ to the non-negative reals, given by
$h(z) = \frac{\partial}{\partial_z} \log(\Gamma(z+1)) + \gamma$, where $\Gamma$ is the gamma function. 

 The following theorems contain formulas for the expectation value of the divergence from the models defined in the previous section, as well as asymptotic
 expressions of these formulas.  The results are based on explicit solutions of the integral~\eqref{integral}, as done
 by Wolpert and Wolf in~\cite{WolpertWolf95:Estimating_functions_of_probability_distributions}.  The proofs are contained in Appendix~\ref{sec:proofs}.

\begin{thm}
  \label{thm:expent}
 If $p\sim\Dir_{\ba}$, then:
  \begin{itemize}
  \item $\< H(p) \> = h(\alpha) -  \sum_{i=1}^N \frac{\alpha_i}{\alpha} h(\alpha_i)$, 
  \item $\< D(p\|u) \> = \log(N) - h(\alpha) + \sum_{i=1}^N \frac{\alpha_i}{\alpha} h(\alpha_i)$. 
  \end{itemize}
In the symmetric case $(\al_1,\ldots,\al_N) =(a,\ldots,a) $, 
  \begin{align*}
    \quad\;\bullet\;\;
    \< H(p) \>
    &=h(Na) -h(a) \\
    & =
    \begin{cases}
      \log(Na)- h(a) + \gamma  + O(1/Na) & \text{ for large $N$ and const. $a$} \\
      \log(N) + O(1/a) & \text{ for large } a\text{ and arb. } N \\
      O(Na) & \text{ as $a\to0$ with bounded $N$ } \\
      h(c) + O(a) & \text{ as $a\to0$ with $Na=c$ }
    \end{cases}
  \end{align*}

  \vspace{-5mm}
  \begin{align*}
  \quad\bullet\;\;
    \< D(p\|u) \>
    &=\log(N) - h(Na) + h(a) \\
    & = 
    \begin{cases}
      h(a) -\log(a) -\gamma +O(1/Na) & \text{ for large $N$ and const. $a$} \\
      O(1/a) & \text{ for large $a$ and arb. $N$ }\\
      \log(N) + O(Na) & \text{ as $a\to0$ with bounded $N$}\\
      \log(N) -h(c) + O(a) & \text{ as $a\to0$ with $Na=c$}.
    \end{cases}
  \end{align*}
\end{thm}
The entropy $H(p)=-\sum_i p_{i} \log p_{i}$ is maximized 
by the uniform distribution~$u$, which satisfies $H(u)=\log(N)$.
For large~$N$, or~$a$, the average entropy is close to the maximum value.  It follows that in these cases the expected
divergence from the uniform distribution~$u$ remains bounded.
The fact that the expected entropy is close to the maximal entropy makes it difficult to estimate the entropy.  
See~\cite{NemenmanShafeeBialek01:Entropy_Inference_Revisited} for a discussion. 

\newcounter{le}

\renewcommand\thele{\arabic{lemma}.\alph{le}}
\refstepcounter{lemma}

\theoremstyle{plain}
\newtheorem{thma}[le]{Theorem}
\begin{thma}
\label{thm:expDpq-p}
  For any $q\in\Delta_{N-1}$, if $p\sim\Dir_{\ba}$, then
  \begin{align*}
    \< D(p\|q) \>_{p} &= \sum\limits_{i=1}^N  \frac{\al_i }{\al}( h(\al_i) - \log(q_i))   -  h(\al) \\ 
    &= D(\tfrac{\ba}{\alpha}\|q) + O(N/\alpha). 
  \end{align*}
  If $\ba=(a,\dots,a)$, then 
  \begin{align*}
    \< D(p\|q) \>_{p} &= D(u\|q) + h(a) +\log(N) - h(Na)  \\
    & = D(u\|q) + (h(a) - \log(a)) - \gamma + O(1/(Na)).
  \end{align*}
\end{thma}

\begin{thma}
\label{thm:expDpq-q}
  For any $p\in\Delta_{N-1}$, if $q\sim\Dir_{\ba}$, then
  \begin{align*}
    \< D(p\|q) \>_{q} &=  \sum_{i=1}^N p_i (\log(p_i) - h(\alpha_i-1)) + h(\alpha-1). \\
    \intertext{If $\alpha_{i}>1$ for all~$i$, then}
    \< D(p\|q) \>_{q} &=D(p\|\tfrac{\ba}{\alpha}) + \sum_{i=1}^{N}O(1/(\alpha_{i}-1)). 
  \end{align*}
\end{thma}
\begin{thma}
  \label{thm:expDpq-pq}
  When $p\sim\Dir_{\ba}$ and $q\sim\Dir_{\tilde\ba}$, then
  \begin{itemize}
  \item[$\bullet$] $\< \sum_{i\in\Xcal}p_{i}\log(q_{i}) \>_{p,q} = \sum_{i=1}^N \frac{\al_i}{\al} h(\tilde\al_i-1) - h(\tilde\al -1)$,
  \item[$\bullet$] $\< D(p\|q) \>_{p,q} = 
-\sum_{i=1}^N \frac{\al_i}{\al} (h(\tilde\al_i-1) - h(\al_i))  +  h(\tilde\al -1) - h(\al) $.
  \end{itemize}
  If $\ba=\tilde\ba$, then
  $\< D(p\|q) \> = \frac{N-1}{\al}$. 
\end{thma}

Consider a sequence of distributions $q^{(N)}\in \Delta_{N-1}$, $N\in\mathbb{N}$. As $N\to\infty$, the expected
divergence $\<D(p\|q^{(N)})\>_p$ with respect to the uniform prior is bounded from above by $1-\gamma+ c$, $c>0$ if and
only if $\limsup_{N\to\infty}D(u\| q^{(N)})\leq c$.
It is easy to see that $D(u\|q)\le c$ whenever $q$ satisfies $q_{x}\ge\frac1Ne^{-c}$ for all $x\in\Xcal$.  Therefore,
the expected divergence is unbounded as $N$ tends to infinity only if the sequence $q^{(N)}$ accumulates at the boundary of the probability simplex. 
In fact, $\lim_{N\to\infty}\<D(p\|q^{(N)})\>\leq 1-\gamma +c$ whenever $q^{(N)}$ is in the subsimplex
$\operatorname{conv}\{(1-e^{-c}) \delta_x+ e^{-c} u \}_{x\in\Xcal}$.  The relative Lebesgue volume of this subsimplex in
$\Delta_{N-1}$ is $(1-e^{-c})^{N-1}$.

For arbitrary Dirchlet priors $\ba^{(N)}$ (depending on~$N$), the expectation value $\<D(p\|q^{(N)})\>_p$ remains
bounded in the limit $N\to\infty$ 
if $D(\tfrac{\ba^{(N)}}{\alpha^{(N)}}\|q^{(N)})$ remains bounded and if $\alpha_i^{(N)}$ is bounded from below by a
positive constant for all~$i$.

If $p,q\sim\Dir_{\ba^{(N)}}$, then the expected divergence $\<D(p\|q)\>_{p,q}$ remains bounded in the limit
$N\to\infty$, provided $\frac{\al^{(N)}}{N}$ is bounded from below by a positive constant.

\begin{thm}
  \label{thm:expMI}
  For a system of $n$ random variables $X_{1},\dots,X_{n}$ with joint probability distribution~$p$, if
  $p\sim\Dir_{\ba}$, then 
\begin{itemize}
  \item $\< H(X_{k}) \> =  h(\alpha)  - \sum_{j=1}^{N_k} \frac{\al^k_j }{\alpha} h(\al^k_j)$,
  \item $\< MI(X_{1},\dots,X_{n}) \> = %
    (n-1)h(\al) +\sum\limits_{i=1}^N \frac{\al_i}{\al} h(\al_i) - \sum\limits_{k=1}^n  \sum\limits_{j=1}^{N_k} \frac{\al^k_j  }{\al} h(\al^k_j)$.
  \end{itemize}
In the symmetric case $(\alpha_1,\ldots,\alpha_N)=(a,\ldots,a)$, 
\begin{itemize}
  \item $\< H(X_{k}) \> =  h(N a)  - h(\frac{N}{N_k} a)$,
  \item $\< MI(X_{1},\dots,X_{n}) \> = (n-1) h(Na) + h(a) - \sum_{k=1}^n h(\frac{N}{N_k} a)$.
  \end{itemize}
If, moreover, $Na/N_{k}$ is large for all $k$ (this happens, for example, when $a$ remains bounded from below by some 
$\varepsilon>0$ and {\em (i)} all $N_{k}$ become large, or {\em (ii)} all $N_{k}$ are  bounded and $n$ becomes large),
then
  \begin{itemize}
  \item $\< H(X_{k}) \> = \log(N_{k}) + O(N_{k}/Na)$,
  \item $\< MI(X_{1},\dots,X_{n}) \> = h(a)-\log(a)-\gamma + O(n \max_{k}N_{k}/Na)$.
  \end{itemize}
\end{thm}
If $Na/N_{k}$ is large for all $k$, then the expected entropy of a subsystem is also close to its maximum, and hence the
expected multi-information is bounded.  This follows also from the fact that the independence model contains the uniform
distribution, and hence $D(p\|\indmod)\le D(p\|u)$.

\begin{thm}
  \label{thm:convmodels}
  Let $\varrho=\{A_{1},\dots,A_{K}\}$ be a partition of $\Xcal$ into sets of cardinalities $|A_{k}|=L_{k}$,
  and let $\nu$ be a reference measure on $\Xcal$.  If $p\sim\Dir_{\boldsymbol\alpha}$, then
  \begin{align*}
    \< D(p\|\Mcal_{\varrho,\nu}) \>    &=  
\sum_{i=1}^N\frac{\alpha_i}{\alpha}(h(\alpha_i) - \log(\nu_i))   - \sum_{k=1}^K \frac{\alpha^\varrho_k}{\alpha} (h(\alpha^\varrho_k) - \log(\nu(A_k))) ,
  \end{align*}
where $\alpha^\varrho_k=\sum_{i\in A_k}\alpha_i$. 
If $\ba=(a,\ldots,a)$,  and (wlog) $\nu(A_k)=L_k/N$,  
  \begin{align*}
   \< D(p\|\Mcal_{\varrho,\nu}) \>    &=  h(a)     - \sum_{k=1}^K \frac{L_k}{N} (h(L_ka)- \log(L_k) )  + D(u\| \nu). 
 \end{align*} 
  If furthermore $N\gg K$, then
  \begin{equation*}
    \< D(p\|\Mcal_{\varrho,\nu}) \>
    = h(a) -\log(a) - \gamma + D(u\|\nu) + O(1/N).
  \end{equation*}
\end{thm}
If $\nu=u$, then $\Mcal_{\varrho,\nu}$ is a partition model and contains the uniform distribution. 
Therefore, the expected divergence is again bounded.  In contrast, the maximal divergence is
$\max_{p\in\Delta_{N-1}}D(p\|\Mcal_\varrho)=\max_k\log(N_k)$. 
The result for mixtures of product distributions of disjoint supports is similar:

\begin{thm}
  \label{thm:cubmix}
Let $\Xcal=\Xcal_1\times\cdots\times\Xcal_n$ be the joint state space of $n$ variables, $|\Xcal|=N$, $|\Xcal_k|=N_k$. 
Let $\varrho=\{A_{1},\dots,A_{K}\}$ be a partition of $\Xcal$ into $K$ support sets $A_k=\Xcal_{1,k}\times\cdots\times\Xcal_{n,k}$, $k=1,\ldots,K$ of the independence model, 
and let $\Mcal_{1}^{\varrho}$ be the model containing all mixtures of $K$ product distributions $p^{(1)},\dots,p^{(K)}$ 
with $\supp(p^{(k)})\subseteq A_{k}$.
\begin{itemize}
\item 
If $p\sim\Dir_{\ba}$, then 
  \begin{multline*}
    \< D(p\|\Mcal_{1}^{\varrho}) \>
    =  \sum_{i=1}^N \frac{\alpha_i}{\alpha}(h(\alpha_i) - h(\alpha)) + \sum_{k=1}^K(|G_k|-1) \frac{\alpha^\varrho_k}{\alpha}(h(\alpha^\varrho_k) - h(\alpha)) \\
-\sum_{k=1}^K \sum_{j\in G_k} \sum_{x_j\in\Xcal_{j,k}} \frac{\alpha^{k,x_j}}{\alpha} (h(\alpha^{k,x_j}) -h(\alpha) ),
  \end{multline*}
  where $\alpha^\varrho_k=\sum_{x\in A_k} \alpha_x$, $\alpha^{k,x_j}=\sum_{y\in A_k\colon y_j = x_j} \alpha_y$,
  and $G_k\subset[n]$ is the set of variables that take more than one value in the block $A_k$.

\item
  Assume that the system is homogeneous, $|\Xcal_i|=N_1$ for all $i$, and that $A_k$ is a cylinder set
  of cardinality $|A_k|=N_1^{m_k}$, $m_{k}=|G_{k}|$, for all $k$.  If
  $(\alpha_1,\ldots,\alpha_N)=(a,\ldots,a)$, then 
  \begin{equation*}
    \< D(p\|\Mcal_{1}^{\varrho}) \>=  
 h(a)  + \sum_{k=1}^K {N_1^{m_k-n}}  ( (m_k-1) h(N_1^{m_k} a)  -m_k h(N_1^{m_k-1}  a) ).
  \end{equation*}

\item 
If $\frac{N_1^{m_k-1}a}{m_k}$ is large for all $k$, then 
  \begin{equation*}
    \< D(p\|\Mcal_{1}^{\varrho}) \> =
h(a) - \log(a) -\gamma
+ O\big(\max_k\frac{m_k}{N_1^{m_k-1}a}\big).
  \end{equation*}
\end{itemize}
\end{thm}

The $k$-mixture of binary product distributions with disjoint supports is a submodel of
the restricted Boltzmann machine model with $k-1$ hidden units, see~\cite{NIPS2011_0307}.  
Hence Theorem~\ref{thm:cubmix} also gives bounds for the expected divergence from restricted Boltzmann machines.

\begin{thm}
  \label{thm:decomposable}
  Consider a decomposable model $\Mcal_{\Scal}$ with junction tree $(V,E)$. If $p\sim\Dir_{\ba}$, then
\begin{multline*}
\<D(p\|\Mcal_{\Scal})\> = 
-\sum_{S\in V} \sum_{j\in\Xcal_S}\frac{\alpha^S_j}{\alpha }h(\al^S_j)  + \sum_{S\in E} \sum_{j\in\Xcal_S} \frac{\al^S_j}{\al} h(\al^S_j) \\
 +(|V|-|E|-1)h(\al) + \sum_{i=1}^N \frac{\al_i}{\al} h(\al_i),
\end{multline*}
where $\al^S_j=\sum_{x\colon x_S=j}\al_x$ for $j\in\Xcal_S$. 
If $p$ is drawn uniformly at random, then 
  \begin{equation*}
    \< D(p\|\Mcal_{\Scal}) \> = \sum_{S\in V} (h(N) - h(N/N_{S})) - \sum_{S\in E} (h(N) - h(N/N_{S})) - h(N) + 1.
  \end{equation*}
  If $N/N_{S}$ is large for all $S\in V\cup E$, then
  \begin{equation*}
    \< D(p\|\Mcal_{\Scal}) \> 
    = 1 - \gamma + O\big(\max_S N/N_S \big).
  \end{equation*}
\end{thm}

\section{Discussion}
\label{sec:interpretation}

In the previous section we have shown that the values of $\< D(p\|\Mcal) \>$ are very similar for different models
$\Mcal$ in the limit of large~$N$, provided the Dirichlet concentration parameters $\alpha_{i}$ remain bounded and the model remains
 small. In particular, if $\alpha_{i}=1$ for all~$i$, then $\< D(p\|\Mcal) \> \approx 1-\gamma$ for large $N$ holds for 
$\Mcal=\{u\}$, for independence models, for decomposable models, for partition models, and for mixtures of product
distributions on disjoint supports (for reasonable values of the hyperparameters $N_{k}$ and $L_{k}$).  Some of these
models are contained in each other, but nevertheless, the expected divergences do not differ much.  The general
phenomenon seems to be the following:
\begin{itemize}
\item If $N$ is large and if $\Mcal\subset\Delta_{N-1}$ is low-dimensional, then the expected divergence is $\<
  D(p\|\Mcal) \> \approx 1-\gamma$, when $p$ is uniformly distributed on~$\Delta_{N-1}$.
\end{itemize}
Of course, this is not a mathematical statement, because it is easy to construct counter-examples: Space-filling curves
can be used to construct one-dimensional models with an arbitrarily low value of $\<D(p\|\Mcal) \>$ (for arbitrary~$N$). 
However, we expect that the statement is true for most models that appear in
practice.  In particular, we conjecture that the statement is true for restricted Boltzmann machines. 

\medskip

In Theorem~\ref{thm:convmodels}, if $\alpha=(a,\dots,a)$, then the expected divergence from a convex exponential family $\Mcal_{\varrho,\nu}$ is
minimal, if and only if $\nu=u$.  In this case $\Mcal_{\varrho,\nu}$ is a partition model.  We conjecture that partition
models are optimal among all (closures of) exponential families in the following sense:
\begin{itemize}
\item For any exponential family $\Ecal$ there is a partition model $\Mcal$ of the same dimension such that $\<
  D(p\|\Ecal) \> \ge \< D(p\|\Mcal) \>$, when $p\sim\Dir_{(a,\dots,a)}$.
\end{itemize}
The statement is of course true for zero-dimensional exponential families, which consist of a single
distribution.  The conjecture is related to the following conjecture from~\cite{Rauh13:Optimal_Expfams}:
\begin{itemize}
\item For any exponential family $\Ecal$ there is a partition model $\Mcal$ of the same dimension such that
  $\max_{p\in\Delta_{N-1}}D(p\|\Ecal) \ge \max_{p\in\Delta_{N-1}}D(p\|\Mcal)$.
\end{itemize}

\appendix

\subsection*{Computations}

Our findings may be biased by the fact that all models treated in Section~\ref{sec:evalues} are exponential families. 
As a slight generalization we did computer experiments with a family of models which are not
exponential families, but unions of exponential families.

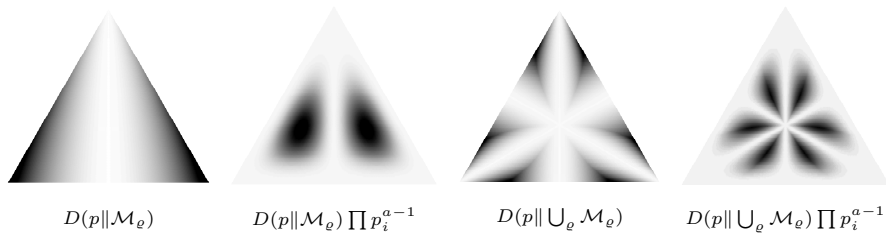
\begin{figure}
\setlength{\unitlength}{.6cm}
\begin{picture}(14,5.4)(-.1,-.9)
\put(0,0){\includegraphics[clip=true,trim=5.5cm 11cm 5cm 5cm,width=5\unitlength]{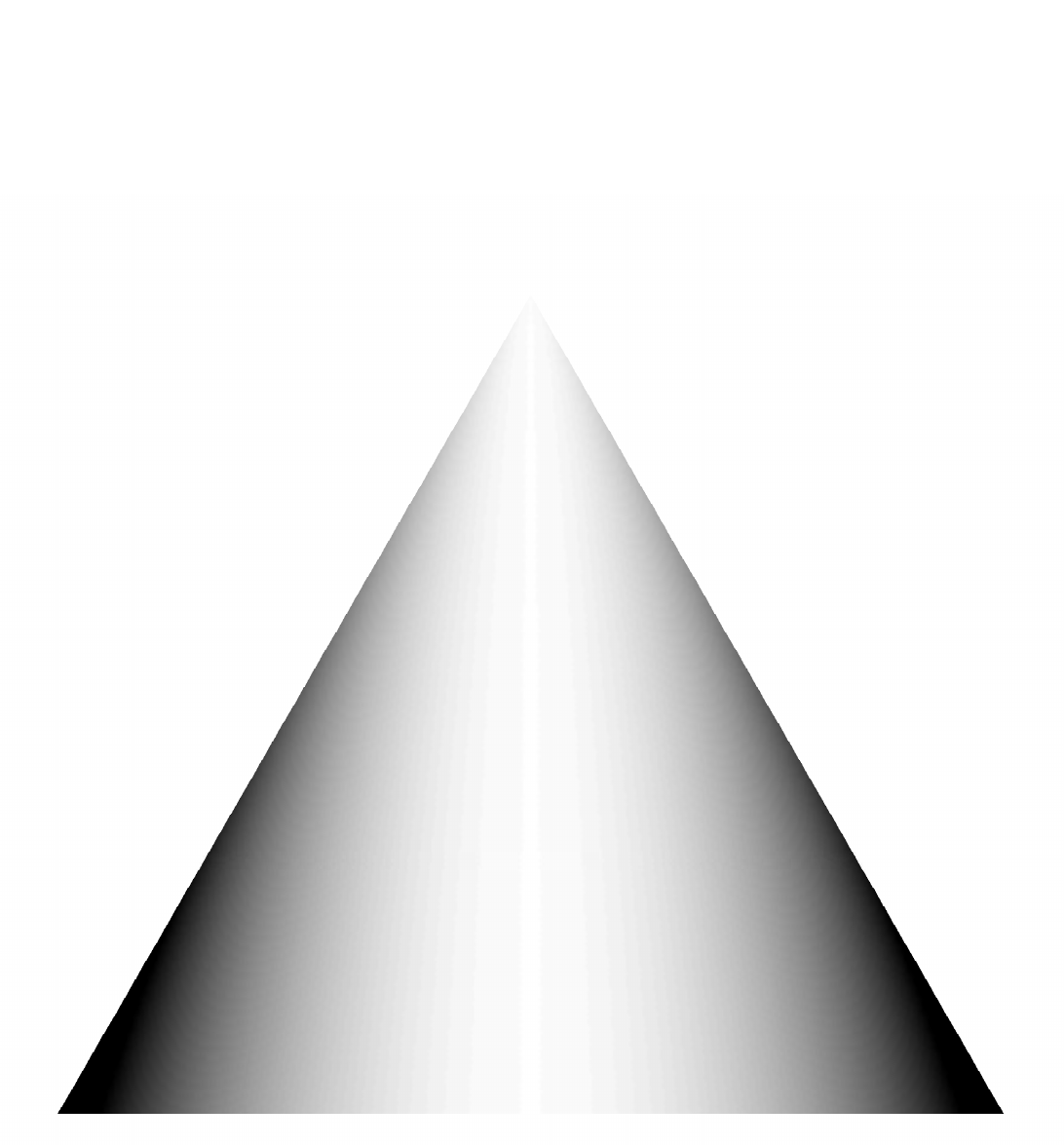}}
\put(5,0){\includegraphics[clip=true,trim=5.5cm 11cm 5cm 5cm,width=5\unitlength]{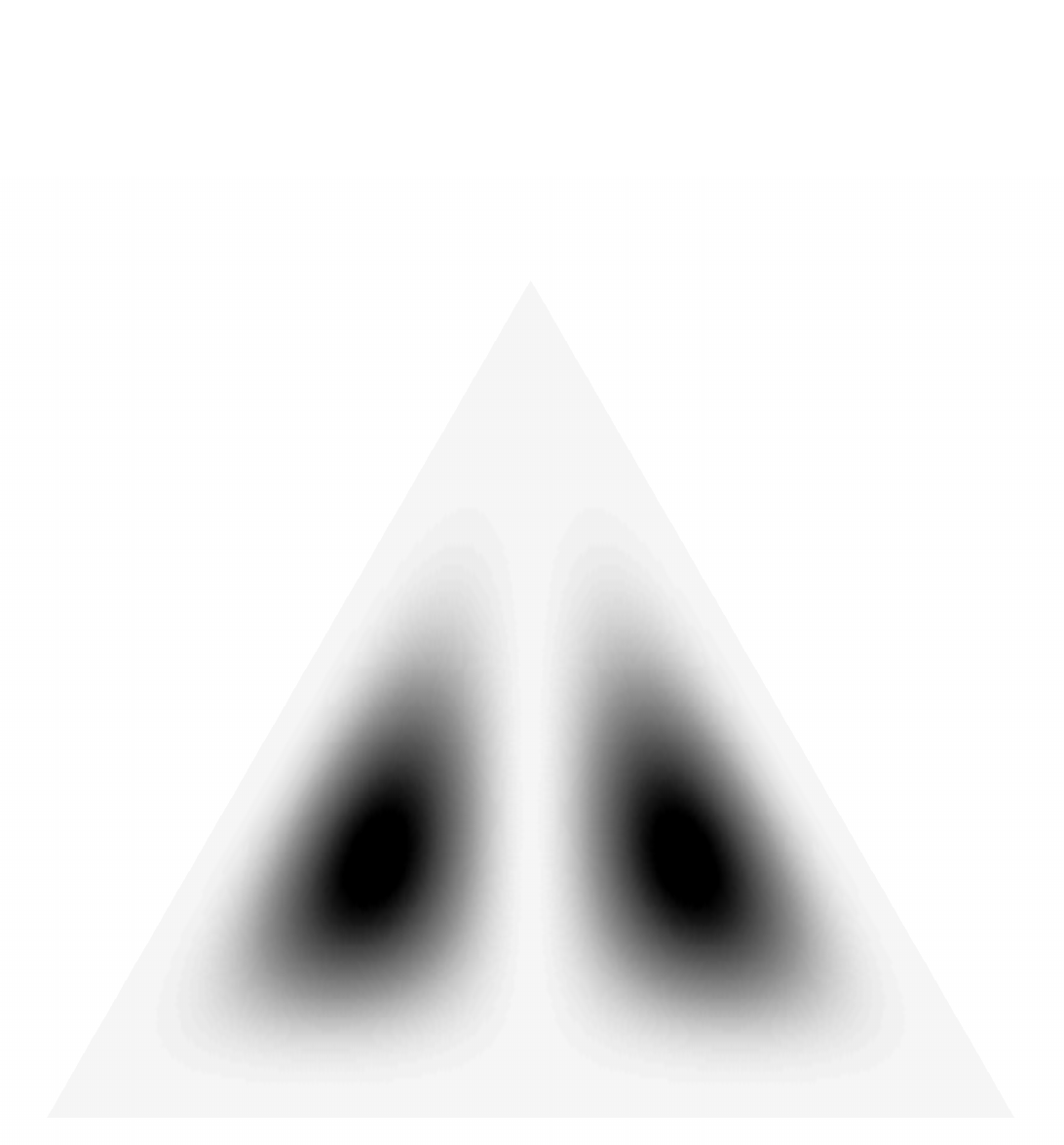}}
\put(10,0){\includegraphics[clip=true,trim=5.5cm 11cm 5cm 5cm,width=5\unitlength]{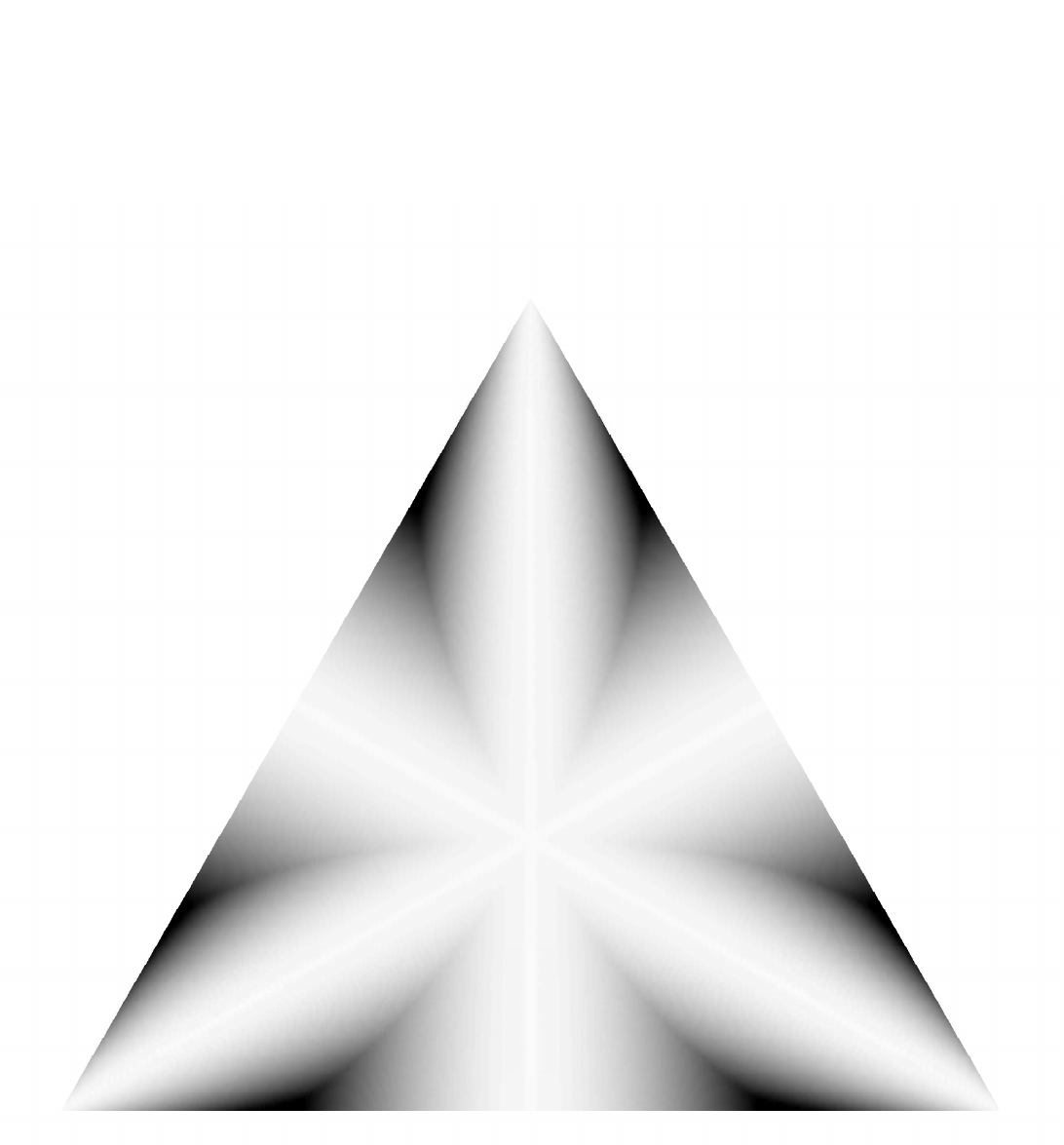}}
\put(15,0){\includegraphics[clip=true,trim=5.5cm 11cm 5cm 5cm,width=5\unitlength]{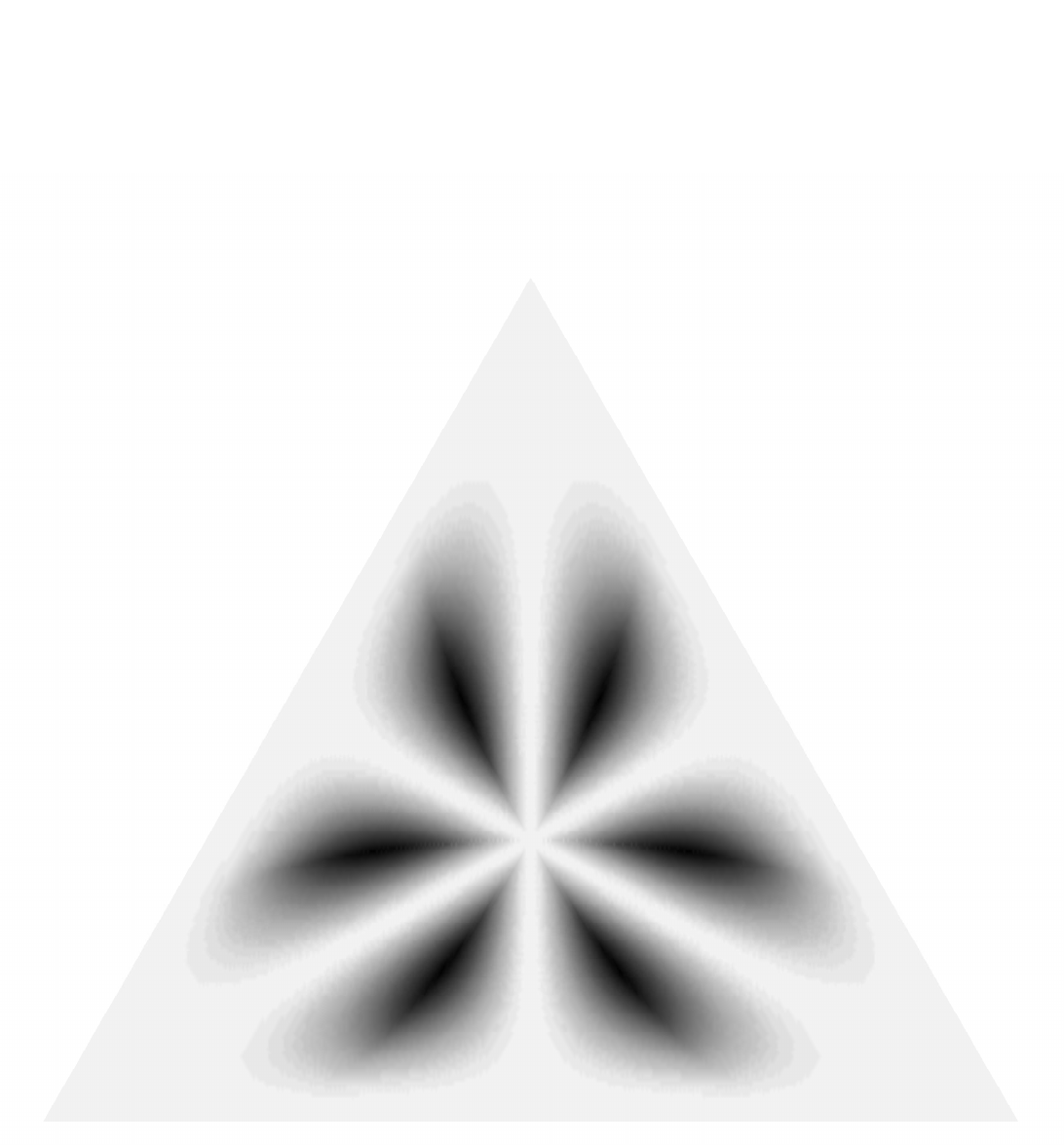}}
\put(0,-.8){\begin{minipage}{5\unitlength}\center\scriptsize$D(p\|\Mcal_\varrho)$\end{minipage}}
\put(5,-.8){\begin{minipage}{5\unitlength}\center\scriptsize$D(p\|\Mcal_\varrho)\prod p_i^{a-1}$\end{minipage}}
\put(10,-.8){\begin{minipage}{5\unitlength}\center\scriptsize$D(p\|\bigcup_\varrho\Mcal_\varrho)$\end{minipage}}
\put(15,-.8){\begin{minipage}{5\unitlength}\center\scriptsize$D(p\| \bigcup_\varrho\Mcal_\varrho)\prod p_i^{a-1}$\end{minipage}}
\end{picture}
\caption{From left to right: Divergence to a partition model with two blocks on $\Xcal=\{1,2,3\}$. Same, multiplied by a symmetric Dirichlet density with parameter $a=5$.  Divergence to the union of the three partition models with two blocks on $\Xcal=\{1,2,3\}$. Same, multiplied by the symmetric Dirichlet density with $a=5$. The shading is scaled on each image individually. 
}
\label{fig}
\end{figure}

Let $\Upsilon$ be a family of partitions of $\{1,\ldots,N\}$, and let $\Mcal_{\Upsilon}=\bigcup_{\varrho\in\Upsilon}\Mcal_{\varrho}$ be the union
of the corresponding partition models.  We are interested in these models, because they can be used to study more difficult models, like restricted Boltzmann machines and deep belief
networks.  Figure~\ref{fig} compares a single partition model on three states with the union of all partition models. 

\begin{figure}
\setlength{\unitlength}{.6cm}
\begin{picture}(6.,7.4)(0.3,-.3)
\put(0,0){\includegraphics[clip=true,trim=4.5cm 9cm 5.6cm 8cm,width=4.3cm]{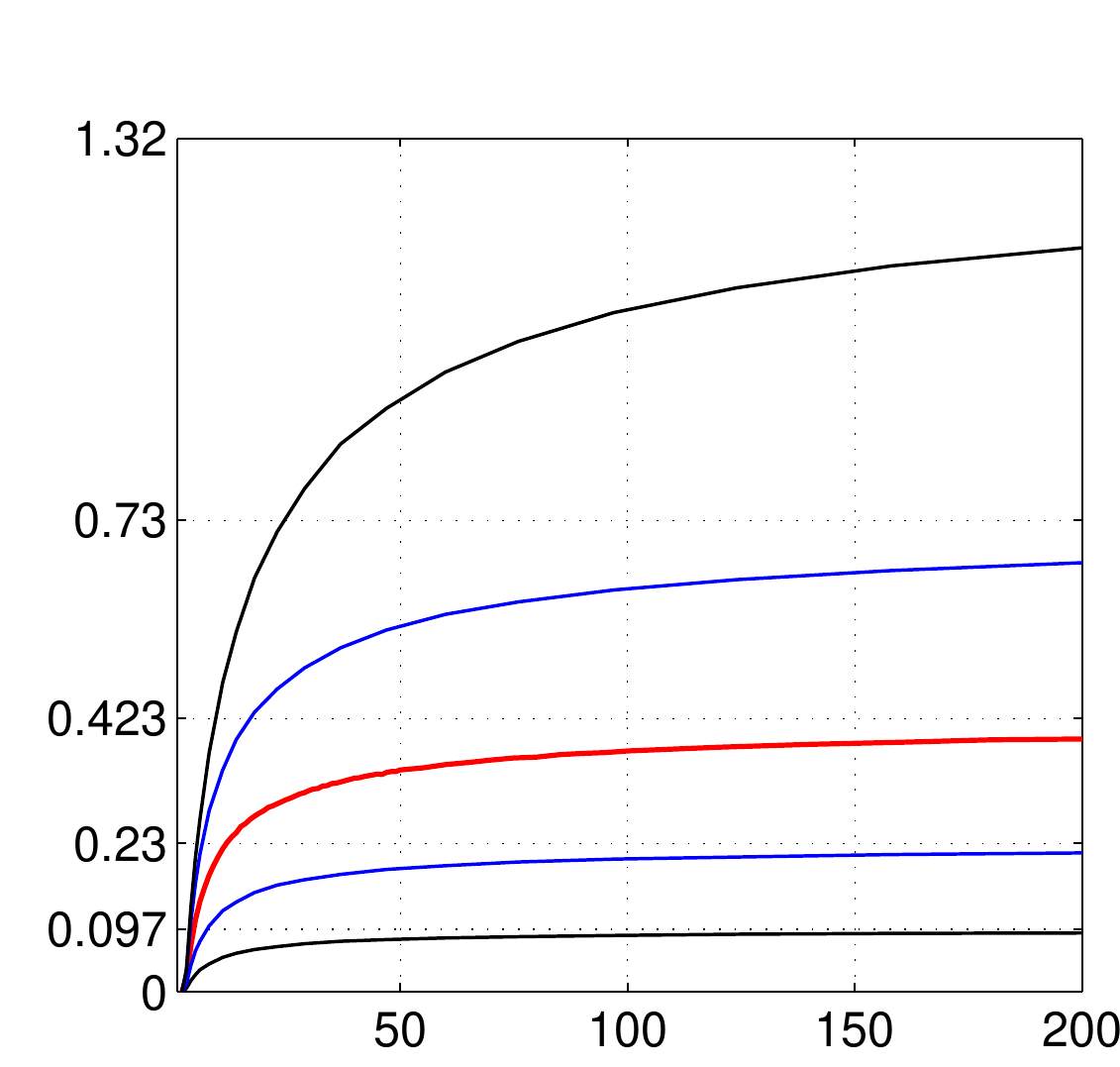}}
\put(3.8,-.5){\small$N$}
\put(4.8,.922){\small$5$}
\put(4.6,1.45){\small$2$}
\put(4.4,2.22){\small$1$}
\put(4.2,3.45){\small$\tfrac12$}
\put(4,5.15){\small$\tfrac15$}
\put(3,5.15){\small$a=$}
\put(1.3,5.2){\fbox{\small$\Upsilon_1$}}
\end{picture}
\setlength{\unitlength}{.6cm}
\begin{picture}(6.3,7.4)(-.1,-.3)
\put(0,0){\includegraphics[clip=true,trim=4.5cm 9cm 5.6cm 8cm,width=4.3cm]{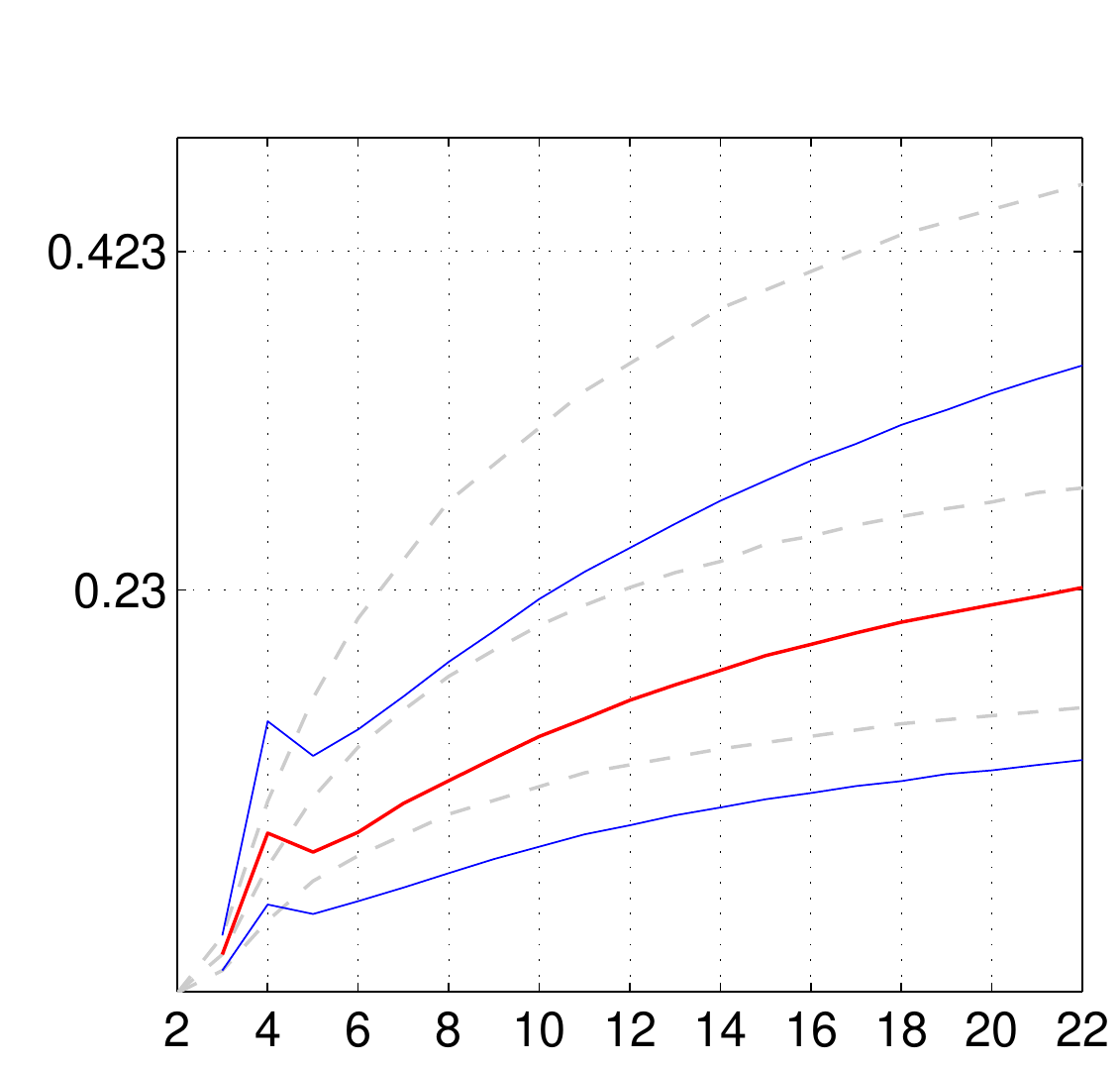}}
\put(3.8,-.5){\small$N$}
\put(5,1.8){\small$2$}
\put(4.6,2.65){\small$1$}
\put(3.2,3.7){\small$a=\tfrac12$}
\put(1.3,5.2){\fbox{\small$\Upsilon_2$}}
\put(1.5,3.2){\textcolor{gray}{\small$\Upsilon_1$}}
\end{picture}
\setlength{\unitlength}{.6cm}
\begin{picture}(6.3,7.4)(-.1,-.3)
\put(0,0){\includegraphics[clip=true,trim=5.9cm 9.65cm 6.25cm 7.8cm,width=4.4cm,height=4.5cm]{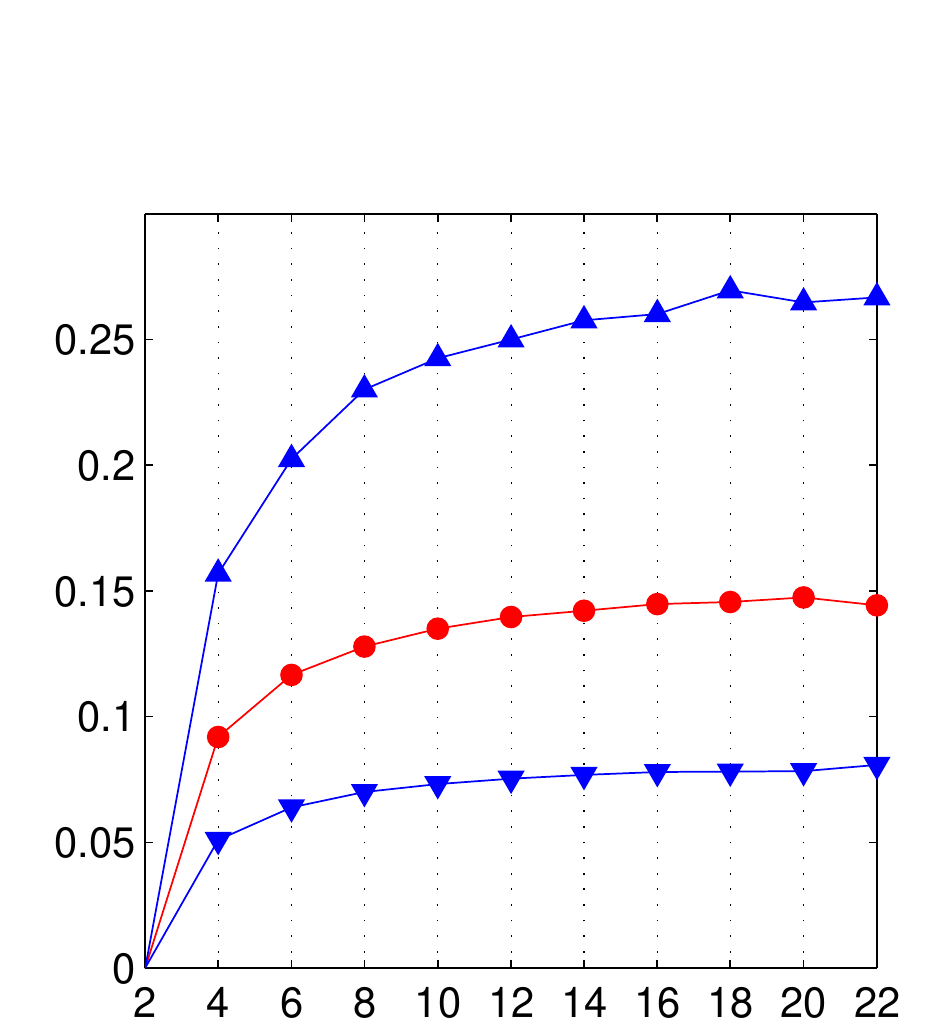}}
\put(3.8,-.5){\small$N$}
\put(5.7,1.4){\small$2$}
\put(5.4,2.55){\small$1$}
\put(4.2,4.5){\small$a=\tfrac12$}
\put(1.3,5.2){\fbox{\small$\Upsilon_{N/2}$}}
\end{picture}
\caption{Expected divergence (numerically) from $\Mcal_{\Upsilon_{k}}$ with respect to 
$\Dir_{(a,\dots,a)}$, for different system sizes $N$ and values of~$a$. 
Left: The case $k=1$. 
The y-ticks are located at $h(a)-\log(a)-\gamma$, which are the limits of the expected divergence from single bipartition models, see  Theorem~\ref{thm:convmodels}. 
Middle: The case $k=2$. 
The peak at $N=4$ emerges, because in this case there are only $3$ different partitions, instead of $\binom{4}{2}$. 
The dashed plot indicates corresponding results from the left figure. 
Right: 
The expected divergence to the union of all $\binom N { N/2}/2$ bipartition models with two blocks of cardinalities $N/2$, for even $N$. %
}
\label{expecteddivunionpartitions}
\end{figure}
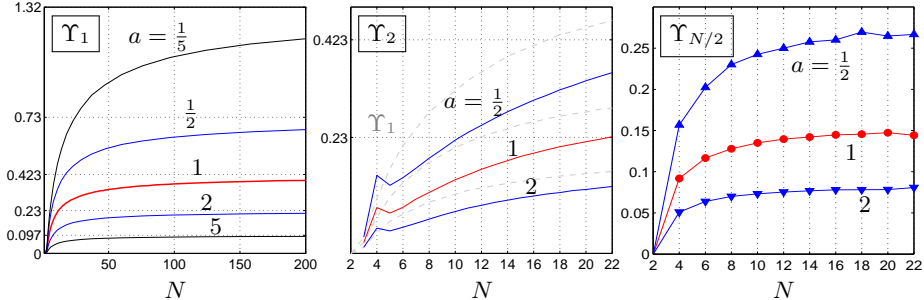

For a given $N$ and $0\le k\le N/2$ let $\Upsilon_{k}$ be the set of all partitions of $\{1,\dots,N\}$ into two blocks of cardinalities
$k$ and $N-k$. 
For different values of $a$ and~$N$ we computed $D(p\|\Mcal_{\Upsilon_{1}})$ for $10\,000$ distributions sampled from $\Dir_{(a,\dots,a)}$, $D(p\|\Mcal_{\Upsilon_{2}})$ for $20\,000$ distributions sampled from $\Dir_{(a,\dots,a)}$, $D(p\|\Mcal_{\Upsilon_{N/2}})$ for $10\,000$--$20\,000$ distributions sampled from $\Dir_{(a,\dots,a)}$ (for $N=22$ only $500$ samples; in this case there are as many as $|\Upsilon_{N/2}|=352\,716$ homogeneous bipartitions). 
The results are shown in Figure~\ref{expecteddivunionpartitions}. 

In the first two cases the expected divergence seems to tend to the asymptotic value of $\<D(p\|u)\>$. 
Observe that $\<D(p\|\Mcal_{\Upsilon_{1}})\> \ge \<D(p\|\Mcal_{\Upsilon_{2}})\>$, unless $N=4$.  
Intuitively this makes sense for two reasons: First, for $\varrho_{1}\in\Upsilon_{1}$ and $\varrho_{2}\in\Upsilon_{2}$, using Theorem~\ref{thm:convmodels} one can show that $\<D(p\|\Mcal_{\varrho_{1}})\> \ge \<D(p\|\Mcal_{\varrho_{2}})\>$; and second, the cardinality of $\Upsilon_{2}$ is much larger than the cardinality of $\Upsilon_{1}$ if $N\ge 4$. 
For small values of $N$ this intuition may not always be correct. 
For example, for $N=8$, the expected divergence from $\Mcal_{\Upsilon_{N/2}}$ is larger than the one from $\Mcal_{\Upsilon_2}$, although in this case $|\Upsilon_{N/2}|=35$ and $|\Upsilon_2|=28$, see Figure~\ref{expecteddivunionpartitions} right. 

We expect that, for large~$N$, it is possible to make $\<D(p\|\Mcal_{\Upsilon_{k}})\>$ much smaller than $\<D(p\|u)\>$
by choosing $k\approx N/2$.  In this case, the model $\Mcal_{\Upsilon_{k}}$ has (Hausdorff) dimension only one, but it
is a union of exponentially many one-dimensional exponential families.

\section{Proofs}
\label{sec:proofs}

The analytic formulas in Theorem~\ref{thm:expent}
are~\cite[Theorem~7]{WolpertWolf95:Estimating_functions_of_probability_distributions}.  The asymptotic expansions are
direct. 

The proof of Theorem~\ref{thm:expDpq-p} makes use of the following Lemma, which is a consequence of~\cite[Theorem~5]{WolpertWolf95:Estimating_functions_of_probability_distributions} and the aggregation property of the Dirichlet distribution:

\begin{lemma}\label{lemma1}
  Let $\{A_1,\ldots,A_K\}$ be a partition of $\Xcal=\{1,\ldots,N\}$, let $\alpha_1,\ldots,\alpha_N$ be positive
  real numbers, and let $\alpha^k=\sum_{i\in A_k}\alpha_i$ for $k=1,\ldots,K$. Then
\begin{align*}
\int_{\Delta_{N\!-\!1}}\!\!\big(\sum_{i\in A_k}p_i \big) \log\big(\sum_{i\in A_k}p_i\big) \prod_{i=1}^N p_i^{\alpha_i-1} \,\rd p =& 
\int_{\Delta_{K\!-\!1}}\!\! p_k^\ast  \log (p_k^\ast) \prod_{k'=1}^K( p^\ast_{k'})^{\alpha^{k'}-1} \,\rd p^\ast\\
=&
\frac{\alpha^k\prod_{k'=1}^K \Gamma(\alpha^{k'}) }{\Gamma(\alpha+1)}(h(\alpha^k) - h(\alpha)).
\end{align*}
\end{lemma}

Let $n=\sum_{j=1}^N n_i$.  Theorem~\ref{thm:expDpq-p} follows from~\cite[Theorem~3]{WolpertWolf95:Estimating_functions_of_probability_distributions}:  
\begin{multline*}
\int_{\Delta_{N-1}} 
p_i \prod_{j=1}^N p_j^{n_j} \,\rd p \; \Big/  \int_{\Delta_{N-1}} \prod_{j=1}^N p_j^{n_j} \,\rd p\\
=\frac{(n_i+1)\prod_{j=1}^N\Gamma(n_j+1)}{\Gamma(N+n+1)} \Big/ \frac{\prod_{j=1}^N\Gamma(n_j+1)}{\Gamma(N+n)}
= \frac{(n_i + 1)}{(N+n)}, 
\end{multline*}
and $D(p\|q) = - H(p) - \sum_{i=1}^N p_{i}\log(q_{i})$. 
By Lemma~\ref{lemma1},
\begin{align*}
  \int_{\Delta_{N-1}} \log(p_i) \prod_{j=1}^N p_j^{n_j} \,\rd p \,\Big/ \int_{\Delta_{N-1}} \prod_{j=1}^N p_j^{n_j} \,\rd p = h(n_i) - h(N+ n -1),
\end{align*}
and this implies Theorems~\ref{thm:expDpq-q} and~\ref{thm:expDpq-pq}.

Theorem~\ref{thm:expMI} is a corollary to Theorem~\ref{thm:expent}, the aggregation property of the Dirichlet priors and
the formula~\eqref{eq:MI} for the multi-information.  Theorem~\ref{thm:convmodels} follows
from eq.~\eqref{eq:div-from-conv}, and Theorem~\ref{thm:decomposable} follows from eq.~\eqref{eq:decomp-div}.  Similarly,
Theorem~\ref{thm:cubmix} follows from the equality
\begin{align*}
  D(p\|\Mcal_{1}^{\varrho}) = \sum_{i=1}^{K}\sum_{x\in A_{i}} p(x) \log\frac{p(x)p(A_{i})^{n-1}}{\prod_{j=1}^{n}(\sum_{y\in A_{i}: y_{j}=x_{j}}p(y))}, 
\end{align*}
which can be derived as follows: The unique solution $q\in\arginf_{q'\in M_{1}^{\varrho}} D(p\|q')$ satisfies
$p(A_{i})=q(A_{i})$ and $q(\cdot|A_{i})\in\arginf_{q'\in\indmod}D(p(\cdot\|A_{i})\|q')$.

\bigskip
\noindent{\bf Acknowledgment:} J.~Rauh is supported by the VW Foundation.
G. Mont\'ufar is supported in part by DARPA grant FA8650-11-1-7145. 

\bibliographystyle{abbrv}
\bibliography{wupes2012}

\begin{thebibliography}{10}

\bibitem{Ay02:Pragmatic_structuring}
N.~Ay.
\newblock An information-geometric approach to a theory of pragmatic
  structuring.
\newblock {\em Annals of Probability}, 30:416--436, 2002.

\bibitem{DrtonSturmfelsSullivant09:Algebraic_Statistics}
M.~Drton, B.~Sturmfels, and S.~Sullivant.
\newblock {\em Lectures on Algebraic Statistics}, volume~39 of {\em Oberwolfach
  Seminars}.
\newblock Birkh\"{a}user, Basel, first edition, 2009.

\bibitem{DirIntro}
B.~A. Frigyik, A.~Kapila, and M.~R. Gupta.
\newblock Introduction to the {D}irichlet distribution and related processes.
\newblock Technical report, Department of Electrical Engineering University of
  Washington, 2010.

\bibitem{MatusAy03:On_Maximization_of_the_Information_Divergence}
F.~Mat\'u\v{s} and N.~Ay.
\newblock On maximization of the information divergence from an exponential
  family.
\newblock In {\em Proceedings of the WUPES'03}, pages 199--204. University of
  Economics, Prague, 2003.

\bibitem{MatusRauh11:Maximization-ISIT2011}
F.~Mat\'u\v{s} and J.~Rauh.
\newblock Maximization of the information divergence from an exponential family
  and criticality.
\newblock In {\em 2011 IEEE International Symposium on Information Theory
  Proceedings (ISIT2011)}, 2011.

\bibitem{NIPS2011_0307}
G.~Mont\'ufar, J.~Rauh, and N.~Ay.
\newblock Expressive power and approximation errors of restricted {B}oltzmann
  machines.
\newblock In {\em Advances in Neural Information Processing Systems 24 (NIPS
  2011)}, pages 415--423, 2011.
\newblock available at
  \url{http://books.nips.cc/papers/files/nips24/NIPS2011_0307.pdf}.

\bibitem{NemenmanShafeeBialek01:Entropy_Inference_Revisited}
I.~Nemenman, F.~Shafee, and W.~Bialek.
\newblock Entropy and inference, revisited.
\newblock In {\em NIPS}, pages 471--478, 2001.

\bibitem{Rauh11:Thesis}
J.~Rauh.
\newblock {\em Finding the maximizers of the information divergence from an
  exponential family}.
\newblock PhD thesis, Universit\"{a}t Leipzig, 2011.

\bibitem{Rauh13:Optimal_Expfams}
J.~Rauh.
\newblock Optimally approximating exponential families.
\newblock {\em Kybernetika}, 2013.
\newblock accepted. Preprint available at \url{http://arxiv.org/abs/1111.0483}.

\bibitem{WolpertWolf95:Estimating_functions_of_probability_distributions}
D.~Wolpert and D.~Wolf.
\newblock Estimating functions of probability distributions from a finite set
  of samples.
\newblock {\em Physical Review E}, 52(6):6841--6854, 1995.

\end{thebibliography}

\end{document}